\documentclass[letterpaper, 10 pt, conference]{ieeeconf}  
\usepackage[utf8]{inputenc}

\usepackage{graphicx}
\usepackage{float} 
\usepackage{tabularx}

\usepackage{mathtools}
\usepackage{booktabs}
\usepackage[table,xcdraw]{xcolor}
\usepackage{dblfloatfix} 
\usepackage{multirow}
\usepackage{scalerel}
\usepackage{multicol}
\usepackage{adjustbox}
\usepackage{cite}
\makeatletter
\let\NAT@parse\undefined
\makeatother
\usepackage[colorlinks=true, allcolors=blue]{hyperref}

\IEEEoverridecommandlockouts

\makeatletter
\newcommand*\titleheader[1]{\gdef\@titleheader{#1}}
\AtBeginDocument{%
  \let\st@red@title\@title
  \def\@title{%
    \bgroup\normalfont\large\centering\@titleheader\par\egroup
    \vskip2.5em\st@red@title}
}
\makeatother

\title{\LARGE \bf DFR-FastMOT: Detection Failure Resistant Tracker for Fast Multi-Object Tracking Based on Sensor Fusion}
\titleheader{\small© 2023 IEEE.  Personal use of this material is permitted.  Permission from IEEE must be obtained for all other uses, in any current or future media, including reprinting/republishing this material for advertising or promotional purposes, creating new collective works, for resale or redistribution to servers or lists, or reuse of any copyrighted component of this work in other works.}

\author{Mohamed Nagy$^{1}$, Majid Khonji$^1$, Jorge Dias$^1$ and Sajid
Javed$^1$
\thanks{$^1$Khalifa University Center for Autonomous Robotic Systems (KUCARS), Department of Electrical Engineering and Computer Science, Khalifa University, Abu Dhabi, United Arab Emirates}
\thanks{ mohamed.nagy@ieee.org, majid.khonji@ku.ac.ae, jorge.dias@ku.ac.ae, sajid.javed@ku.ac.ae}}

\begin{document}

\maketitle
\thispagestyle{empty}
\pagestyle{empty}

\begin{abstract}
Persistent multi-object tracking (MOT) allows autonomous vehicles to navigate safely in highly dynamic environments. One of the well-known challenges in MOT is object occlusion when an object becomes unobservant for subsequent frames. The current MOT methods store objects information, like objects' trajectory, in internal memory to recover the objects after occlusions. However, they retain short-term memory to save computational time and avoid slowing down the MOT method. As a result, they lose track of objects in some occlusion scenarios, particularly long ones. In this paper, we propose DFR-FastMOT, a light MOT method that uses data from a camera and LiDAR sensors and relies on an algebraic formulation for object association and fusion. The formulation boosts the computational time and permits long-term memory that tackles more occlusion scenarios. Our method shows outstanding tracking performance over recent learning and non-learning benchmarks with about $3\%$ and $4\%$ margin in $MOTA$, respectively. Also, we conduct extensive experiments that simulate occlusion phenomena by employing detectors with various distortion levels. The proposed solution enables superior performance under various distortion levels in detection over current state-of-art methods. Our framework processes about $7,763$ frames in $1.48$ seconds, which is seven times faster than recent benchmarks. The framework will be available at \href{https://github.com/MohamedNagyMostafa/DFR-FastMOT}{https://github.com/MohamedNagyMostafa/DFR-FastMOT}. 
\end{abstract}

\section{Introduction} 

\begin{figure}[t!]
\centering
\includegraphics[width=\linewidth]{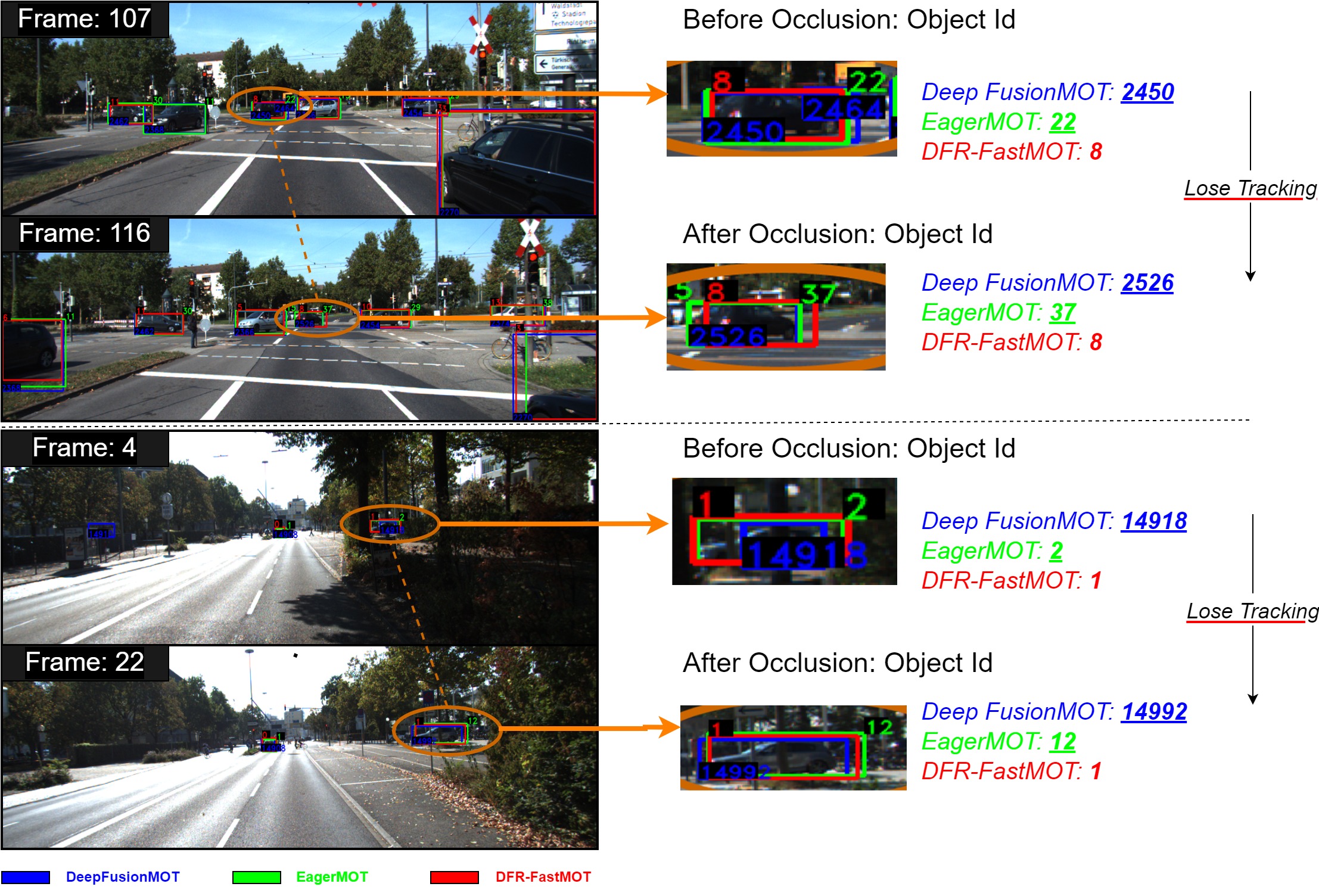}
\caption{\label{fig:comp_live}Our approach recovered objects in two occlusion scenarios when DeepFusionMOT \cite{wang2022deepfusionmot} and Eager-MOT \cite{kim2021eagermot}  methods could not retrieve them.}

\end{figure}

Multi-object tracking (MOT) provides information about the surrounding objects, allowing autonomous vehicles (AVs) to avoid collisions with other cars by making proper navigation decisions. AVs rely on sensors such as cameras and LiDAR to obtain sufficient information for object tracking. We categorize the recent research work into two groups: the first group uses mono-sensor, and the other group employs multi-sensors to collect information for MOT.\\
As shown by Chaabane  \cite{chaabane2021deft}, Tokmako  \cite{tokmakov2021learning}, and Wu  \cite{wu2021tracklet}, they rely on either a camera or LiDAR sensor to localize and track objects. However, sensors usually have limitations; for example, LiDAR performance decreases in fog and sandy weather. On the other hand, a camera sensor resolution drops in night scenes. Accordingly,  Wang \cite{wang2022deepfusionmot} and Kim \cite{kim2021eagermot} fuse data from a camera and LiDAR sensors as well as employ Kalman Filter (KF)  \cite{kf} to track objects by trajectory estimation. Similarly, Wu \cite{wu2021tracklet} fuses LiDAR, IMU, and GPS sensors to track objects in 3D, considering object orientations. Even though Wu \cite{wu2021tracklet} achieves high tracking performance as a learning-based solution compared to other benchmarks, including non-learning based, the tracker skeleton is vast. It requires evolved hardware, like multiple GPUs, which makes the method not applicable to some applications, such as mobile robots. Besides, learning-based solutions fail under prolonged occlusion scenarios when an occluded object does not appear within the frame's stack fed to the deep learning model. Furthermore, the proposed non-learning solutions, like Wang \cite{wang2022deepfusionmot} and Kim \cite{kim2021eagermot}, sacrifice the tracking performance for less computational time by maintaining a short-term memory for objects during occlusions. Thus, these methods fail to capture some occlusions, as shown in Figure \ref{fig:comp_live}.

In this work, we propose an MOT method that tackles the limitations of the previous work, non-learning (Low performance caused by object occlusion) and learning (High computational time) solutions. Our method utilizes data from LiDAR and camera sensors. It relies on an algebraic model to associate and fuse objects, enhancing the computational time and allowing memory expansion to capture more occlusion scenarios. The method can utilize mono or multi-detectors to perform MOT, which makes it applicable to several autonomous applications, like mobile robots.\\

We summarize our contributions as follows:
\begin{enumerate}
    \item We propose a light MOT framework with a remarkable tracking performance over the current learning and non-learning MOT methods with less computational time, which is about seven times faster than recent non-learning methods.
    \item We simulate occlusion phenomena by employing various detection distortion levels to evaluate the performance of our solution under object occlusion. The framework shows superior performance throughout the experiments over other benchmarks. 
    \item We show the framework's capability to track objects under different types of occlusion, such as off-scene objects, objects with single detection information 2D/3D, and multi-object occlusion.
\end{enumerate}

\section{Literature Review}
Many research works propose outstanding performance in object tracking  \cite{shuai2021siammot, zheng2021improving, stadler2021improving, zhou2020tracking, zhang2020multiplex} that employ a tracking-by-detection paradigm. Bewley et al.  \cite{bewley2016simple} introduce SORT for online MOT where they use Kalman filter (KF) \cite{kf} for object's trajectory estimation using historical observations. They operate the Hungarian algorithm  \cite{kuhn1955hungarian} to associate the objects in the subsequent frames. On the other hand, Bochinski et al.  \cite{bochinski2017high} take advantage of high-rate frames and sophisticated detectors to associate objects using Intersection Over Union (IoU). Hence, they achieve a competitive speed as an MOT approach. \\
Despite the outstanding performance achieved by these trackers, object occlusion remains challenging since they need to handle memory for the tracked objects. Moreover, they assume the observant is always static, which is not applicable in AV applications where the observant is usually in motion. In addition, they utilize mono-sensor for MOT when it is preferable to fuse multi-sensor data to overcome mono-sensor faults.\\
Other research works propose MOT solutions for AVs. We will discuss the research work that uses mono-sensor  \cite{chaabane2021deft, reich2021monocular, tokmakov2021learning, wu2021tracklet, luo2020fast} and multi-sensor  \cite{wang2022deepfusionmot, kim2021eagermot, wu20213d, unscent, extdistance} and emphasize the pros and cons of each.

Chaabane et al.  \cite{chaabane2021deft} propose a joint deep learning model that composes detection and tracking tasks relying on object appearance in captured camera frames by integrating an LSTM model  \cite{hochreiter1997long} to capture motion constraints. However, This method needs to tackle the object occlusion problem. Hence, Tokmakov et al.  \cite{tokmakov2021learning} introduce an object permanence tracker supported by a spatio-temporal and recurrent memory module that identifies observed objects' location and identities using whole history. Although they have superior performance for tracking occluded objects, they maintain a permanent history for objects that may harm association efficiency in long-term running. Besides, utilizing one source of information reduces the robustness of the solution.

On the other hand, research works \cite{wu20213d, wang2022deepfusionmot,kim2021eagermot} consider object tracking using multi-sensor fusion. The methods obtain information using sophisticated detectors from 2D  camera frames and 3D LiDAR point cloud. Kim et al.  \cite{kim2021eagermot} use pre-trained deep learning models  \cite{yin2021centerbased, chen2019mmdetection, cai2017cascade} for object detection. Next, they engage KF to estimate object trajectories by associating the detection outcomes with prior observations using IoU. The limitation of this work is that they use a naive KF model that assumes objects always have a constant velocity, which does not apply to objects like cars. Wang  \cite{unscent} and Kim  \cite{extdistance} apply non-linear filters to estimate complex motion for tracked objects. Kim  \cite{extdistance} uses object distance from LiDAR and Radar sensors to track objects utilizing the extended KF and shows the result by operating Prescan simulator  \cite{8479296} in different scenarios. In contrast, Wang  \cite{unscent} proposes a modified version of the unscented KF for state estimation that improves tracking accuracy. Meanwhile, Weng et al.  \cite{weng20203d} use baseline algorithms, Hungarian algorithm  \cite{kuhn1955hungarian}, and show the capability of baseline algorithms to achieve a comparable tracking accuracy to the proposed deep learning solutions. 

The previous work employs a short-term memory for objects that prevents capturing some occlusion scenarios, which eventually influences the overall tracking performance. In this work, we tackle this problem by introducing an algebraic formulation for association and fusion steps that enhances the MOT computational time and allows the integration of long-term memory.


\section{Methodology}

\begin{figure*}[t!]
\centering
\includegraphics[width=\linewidth, height=4cm]{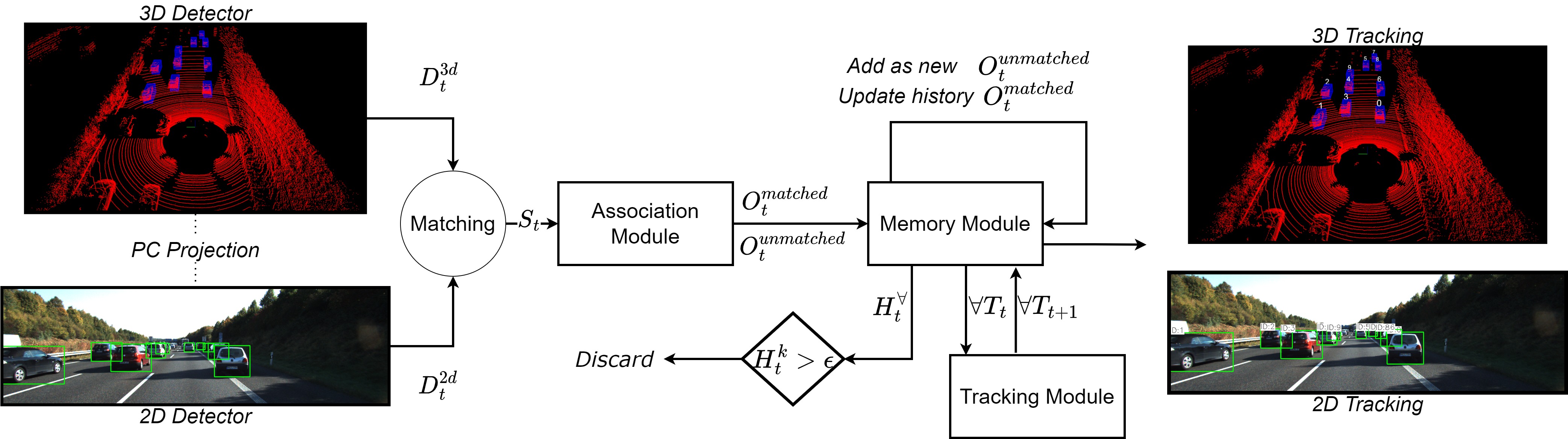}
\caption{\label{fig:overview}\footnotesize A high-level overview of the proposed framework presents a pipeline of the vital modules in the framework. The framework accepts the detection of objects in sensor data and unifies the objects to prevent object duplication. Next, it associates the unified objects with objects stored in the memory. Eventually, the framework updates the trajectory estimation of all objects in the memory and estimates their state for the next frame.}
\end{figure*}

The framework accepts $2D$ and $3D$ detection from camera $D_t^{2d}$ and LiDAR $D_t^{3d}$ at time $t$, and unifies the detected objects in a matching phase to prevent duplicated information of the same object. Next, we associate the detection with prior observed objects in the memory to obtain sets of unmatched objects  $O_t^{unmatched}$ and matched objects $O_t^{matched}$. The memory module updates the history of the matched objects $O_t^{matched}$ and adds the unmatched objects $O_t^{unmatched}$ as new objects. Furthermore, the memory module discards aged objects that did not appear for a number of frames $H_t$ that exceeds $\epsilon$ frames. We eventually employ KF with constant acceleration to update objects' trajectories $T_t^{\forall}$ stored in the memory and obtain state estimation for the subsequent frames based on changes in the trajectories. Figure \ref{fig:overview} shows an overview of the framework.

In this section, we begin by exploring the detection inputs for the framework, Section \ref{subsec:detection}. Next,  we will explain the association and fusion mechanism in Section \ref{subsec:association}. Then, we will discuss the tracking module in Section \ref{subsec:tracking}, followed by the memory management module of the framework in Section \ref{subsec:memory}. 

\subsection{Detection Module}
\label{subsec:detection}
    
The framework requires input data from a camera and LiDAR sensors; however, it may rely solely on 2D or 3D detectors to perform tracking. The framework can employ a mono-detector, either 2D or 3D, and obtain the other detection information using calibration parameters and point cloud projection that allows transformation from camera to LiDAR system coordinates and vice versa, as shown in Figure \ref{fig:overview}. The essence of employing mono-detection is to make the framework applicable for real-time applications when employing two detectors will be costly in terms of power consumption and hardware requirements for some AVs, particularly for mobile robots. 

In the case of multi-detectors, we involve an additional step, matching, where we match detection $D_t^{2d}$ and $D_t^{3d}$ from the detectors to prevent detection duplication for the same object. To accomplish so, we use the same procedures of mono-detector by obtaining $2D$ detection for objects from $D_t^3d$. Then, we consider a detection duplication of an object when the transformed $2D$ bounding box from $D_t^{3d}$ matches one of the objects in $D_t^{2d}$. In the case of matching, we assign the matched data in $D_t^{2d}$ and $D_t^{3d}$ to the object; otherwise, the detection will be classified as two different objects. Hence, the outcome is a set of objects $S_t$ that contains objects with $2D$, $3D$, or a combination of the detection. 

\subsection{Association Module}
\label{subsec:association}
    \begin{figure*}[t!]
\centering
\includegraphics[width=\linewidth]{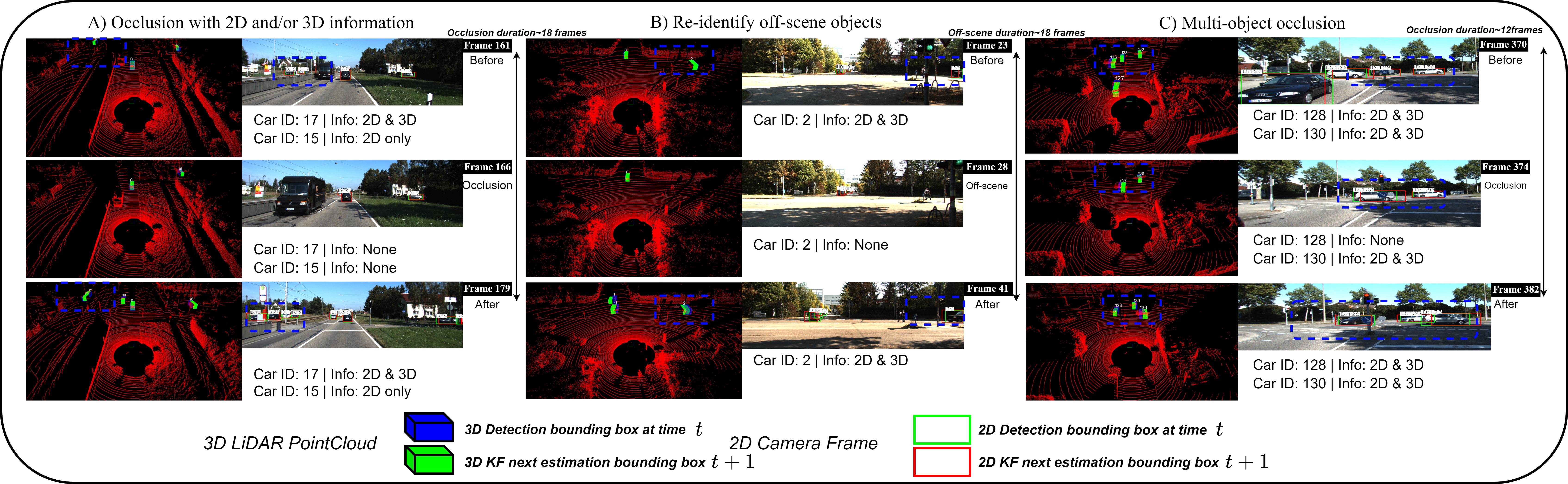}
\caption{\label{fig:occlusion} \footnotesize The performance of the framework over various occlusion scenarios. A) Two occluded cars with either 2D or 3D information. B) Re-identify off-scene objects caused by ego motion. C) Two cars occluded by a third tracked car.}
\end{figure*}

The framework performs association and fusion steps in an algebraic formulation that makes it \textit{\textbf{seven times faster}} than recently published benchmarks \cite{wang2022deepfusionmot} \cite{kim2021eagermot}. We initially introduce an association matrix, Equation \ref{eq:association_matrix}, for each sensor; association matrix $M_c$ for the camera and $M_l$ for the LiDAR sensor. The matrices have the same formulation as Equation \ref{eq:association_matrix}, where the number of rows $m$ is the number of current detected objects by the sensor, and the number of columns $n$ is the number of objects stored in the memory. $v_{ij}$ represents the association value between a recent detected object $i$ and object $j$ stored in the memory. Sections \ref{subsec:camera_ass_matrix} and \ref{subsec:lidar_ass_matrix} describe how the association value is assigned based on the sensor's type.

\begin{equation}
\small
\label{eq:association_matrix}
    M_{m \times n} = \left.\left( 
                  \vphantom{\begin{array}{c}1\\1\\1\\1\\1\end{array}}
                  \smash{\underbrace{
                      \begin{array}{cccc}
                             v_{0,0}&v_{0,1}&v_{0,2}&\cdots\\
                             v_{1,0}&v_{1,1}&v_{1,2}&\cdots\\
                             v_{2,0}&v_{2,1}&v_{2,2}&\cdots\\
                             \vdots&&&\ddots\\
                             v_{m,0}&v_{m,1}&v_{m,2}&\cdots
                      \end{array}
                      }_{ n \text{:prior observed objects}}}
              \right)\right\}
              \,m \text{:observed objects}
\end{equation}    
\newline

In the fusion part, we merge the association matrices using Equation \ref{eq:fusion} to obtain a fused association matrix $M_f$. $\alpha_c$ and $\alpha_l$ represent the significance of the matrices. For example, If LiDAR detection provides a robust association result rather than camera detection, we would trust $M_l$ over $M_c$ by increasing $\alpha_l$ and decreasing $\alpha_c$.

\begin{equation}
\small
\label{eq:fusion}
\begin{split}
   M_f &= \alpha_c M_c + \alpha_l (1 - M_l)\\
   \alpha_c + \alpha_l &= 1\\ 
   \alpha_c, \alpha_l  &\leq 1
\end{split}
\end{equation}

Based on the matching step explained in Section \ref{subsec:detection}, we can guarantee that both $M_c$ and $M_l$ will have the same dimension. In case detection for an object does not exist, we assign the object's row in the matrix by the matrix threshold based on the sensor type, introduced in Sections \ref{subsec:camera_ass_matrix} and \ref{subsec:lidar_ass_matrix}.

\subsubsection{Camera Association Matrix ($M_c$)}
\label{subsec:camera_ass_matrix}
To associate two objects in 2D frames, we use IoU under a certain threshold $a_c$. The association value $v_{ij}$ between two objects can be represented as follows:  

\[
\small
  v_{ij} = \left\{
     \begin{array}{@{}l@{\thinspace}l}
       \text{$v_{IoU}$}  &: \text{$v_{IoU}$}\ge a_c,\\
       \text{0} &: \text{Otherwise.}\\
     \end{array}
   \right.
\]

If $v_{IoU}$ is less than $a_c$, the corresponding $v_{ij}$ will equal $0$, which means the objects are not matched. Otherwise, we maintain $v_{IoU}$ for $v_{ij}$. Accordingly, the $M_c$ matrix shows matched objects with a corresponding $v_{ij}$ close to $1$. The time requires for matrix traversal is $\mathcal{O}(mn)$.\\

\subsubsection{LiDAR Association Matrix ($M_l$)}
\label{subsec:lidar_ass_matrix}
To handle more object occlusions, we do not utilize IoU for the 3D association matrix; we employ $3D$ centroid distance  \cite{object-aware} instead. When a prolonged occlusion happens, it is less likely to have an intersection between the estimated bounding box and the currently detected bounding box. Hence, we compute the Euclidean distance between centroids of the $3D$ bounding boxes, which gives more flexibility to re-identify occluded objects. We perform a similar strategy as the $M_c$ matrix, $\mathcal{O}(mn)$, by having $a_l$ as a Euclidean distance threshold. $v_{ij}$ will be assigned as follows:

\[
\small
  v_{ij} = \left\{
     \begin{array}{@{}l@{\thinspace}l}
       \text{$v_{dist}$}  &: \text{$v_{dist}$} < a_l,\\
       \text{$a_l$} &: \text{Otherwise.}\\
     \end{array}
   \right.
\]

We match the objects when $v_{dist}$ converges to zero; otherwise, we assign the threshold $a_l$ that refers to these two objects as not matched.

To equally fuse $M_c$ and $M_l$, we need to ensure that both matrices have values between $0$ and $1$, which is not the case for $M_l$. Therefore, we normalize $M_l$ by dividing by the maximum value, $a_l$, to obtain a normalized matrix. Eventually, we need to reverse this behavior to match $M_c$ matrix. Thus, we perform $(1 - M_l)$ as shown in Equation \ref{eq:fusion} to obtain a reversed behavior. Thus, $v_{ij}$ represents matched objects when it converges to one instead.

        
     
        

               

Using Equation \ref{eq:fusion}, we obtain a fused matrix $M_f$ that combines the association matrices from the sensors. We eventually perform an algorithm similar to the Hungarian algorithm  \cite{kuhn1955hungarian} to obtain the final association of the objects from $M_f$ under association threshold $a_f$. We pick the maximum value $v_{ij}$ that exists in $M_f$. Then, we associate the corresponding objects and repeat the same algorithm till the maximum value $v_{ij}$ is below the threshold $a_f$. In this case, the algorithm terminates. In the worst case, searching for all elements in $M_f$ takes $\mathcal{O}(m^2n^2)$, which is the complexity of matching all objects in $M_f$. Thus, the overall computational time for the association step of each sensor ($M_c, M_l$) and the fused matrix ($M_f$) is $\mathcal{O}(2mn) + \mathcal{O}(m^2n^2) \rightarrow \mathcal{O}(m^2n^2)$.

\subsection{Tracking Module}
\label{subsec:tracking}
\begin{table*}[!t]

\begin{tabular}{@{}|cccccccccccc|@{}}

     \toprule
      Detector & Method&HOTA$\uparrow$ & MOTA$\uparrow$ &MOTP $\uparrow$ & DetA$\uparrow$ & AssA$\uparrow$ & IDSW$\downarrow$ & IDF1$\uparrow$ & MT$\uparrow$ & ML$\downarrow$ & Frag$\downarrow$  \\ 
     \toprule
     2D YOLOv3 \cite{DBLP:journals/corr/abs-1804-02767} & EagerMOT \cite{kim2021eagermot} &36.5\%&	41.6\%&76.4\%&35.4\%&38\%&792&48.1\%&105&124&855
 \\
      + & DeepFusion-MOT \cite{wang2022deepfusionmot}&30\%&31.8\%&\textbf{77.4\%}&27.4\%&33.1\%&696&40.3\%&28&241&\textbf{819}
\\
     3D PC Projection & \textbf{DFR-FastMOT(Our)}&\textbf{39.2\%}&\textbf{44.5\%}&76.3\%&\textbf{36.3\%}&\textbf{42.8\%}&\textbf{386}&\textbf{53.4\%}&\textbf{113}&\textbf{106}&907
\\
     \midrule

     2D RCC \cite{8099570} & EagerMOT \cite{kim2021eagermot} &70.8\%&82.2\%&\textcolor{red}{\underline{\textbf{90.8\%}}}&79.5\%&63.3\%&1303&74\%&413&26&213
 \\
      + & DeepFusion-MOT \cite{wang2022deepfusionmot}&42.6\%&40.2\%&90.7\%&41.7\%&43.7\%&1203&44.9\%&81&211&1063
\\
     3D PC Projection & \textbf{DFR-FastMOT(Our)}&\textbf{81.9\%}&\textcolor{red}{\underline{\textbf{91\%}}}&90.7\%&\textcolor{red}{\underline{\textbf{83.6\%}}}&\textbf{80.3\%}&\textbf{215}&\textbf{90.1\%}&\textbf{485}&\textbf{12}&\textbf{239}
\\
     \midrule
      2D RCC \cite{8099570} & EagerMOT \cite{kim2021eagermot} &69.1\%&67.2\%&85.2\%&62.6\%&76.5\%&112&80.9\%&499&10&316\\
      + & DeepFusion-MOT\textbf{*} \cite{wang2022deepfusionmot}&77.5\%&87.3\%&86.6\%&75.4\%&79.9\%&\textbf{83}&\textcolor{red}{\underline{\textbf{91.9\%}}}&450&14&301\\
    3D PorintRCNN \cite{Shi_2019_CVPR}& \textbf{DFR-FastMOT(Our)}&\textcolor{red}{\underline{\textbf{82.8\%}}}&\textbf{90.7\%}&\textbf{90.6\%}&\textbf{83.1\%}&\textcolor{red}{\underline{\textbf{82.6\%}}}&177&91.4\%&\textbf{503}&\textbf{8}&\textbf{238}
\\
     \midrule
      2D RCC \cite{8099570} & EagerMOT \cite{kim2021eagermot} &78\%&87.3\%&87.6\%&76.8\%&79.5\%&\textbf{91}&89.3\%&509&7&246\\
      + & DeepFusion-MOT \cite{wang2022deepfusionmot}&77.2\%&85.8\%&86.8\%&74.9\%&79.9\%&136&\textbf{90.9\%}&426&22&280
\\
     3D PointGNN \cite{Point-GNN} & \textbf{DFR-FastMOT(Our)}&\textbf{82.2\%}&\textbf{90.2\%}&\textbf{90.5\%}&\textbf{82.5\%}&\textbf{82\%}&189&90.5\%&\textcolor{red}{\underline{\textbf{516}}}&\textcolor{red}{\underline{\textbf{6}}}&\textbf{224}

\\
\midrule
2D TrackRCNN \cite{Voigtlaender19CVPR_MOTS} & EagerMOT \cite{kim2021eagermot} &68.6\%&63.2\%&85\%&60.8\%&\textbf{77.6\%}&102&80.4\%&\textbf{508}&\textbf{10}&\textbf{262}
 \\
      + & DeepFusion-MOT \cite{wang2022deepfusionmot}&62.2\%&57.7\%&\textbf{86.9\%}&51.5\%&75.3\%&\textcolor{red}{\underline{\textbf{65}}}&73.2\%&307&129&286
\\
     3D PorintRCNN \cite{Shi_2019_CVPR}& \textbf{DFR-FastMOT(Our)}&\textbf{70\%}&\textbf{80.1\%}&82.6\%&\textbf{67.7\%}&72.7\%&193&\textbf{85.9\%}&425&18&608
\\
     \midrule

     2D TrackRCNN \cite{Voigtlaender19CVPR_MOTS} & EagerMOT \cite{kim2021eagermot} &\textbf{78.2\%}&\textbf{86\%}&\textbf{87.5\%}&\textbf{75.8\%}&\textbf{80.8\%}&\textbf{83}&\textbf{90.2\%}&\textbf{522}&\textcolor{red}{\underline{\textbf{6}}}&\textcolor{red}{\underline{\textbf{184}}}
 \\
      + & DeepFusion-MOT \cite{wang2022deepfusionmot}&66\%&65.9\%&86.2\%&58.6\%&74.6\%&133&79  \%&351&86&386
\\
     3D PointGNN \cite{Point-GNN} & \textbf{DFR-FastMOT(Our)}&69.7\%&80\%&82.6\%&67.7\%&72.1\%&203&85.2\%&429&13&561
\\
    \bottomrule
\end{tabular}
\caption{\footnotesize \label{tab:detection_eval}Results using the KITTI evaluation tool on the training and evaluation dataset employing three detector performances \textit{(Poor/Moderate/High)} with two of the recent benchmarks, EagerMot \cite{kim2021eagermot} and DeepFusion-MOT \cite{wang2022deepfusionmot}. \textbf{Black bold} highlights are the highest achieved value per metric for each detector. \textcolor{red}{\underline{\textbf{Red bold}}} highlights are the highest achieved value per metric over all detectors. We provide this experiment results to the public so they can access the results on individual streams of the dataset throughout this link:  \href{https://github.com/MohamedNagyMostafa/KITTI-MOT.Bench-Evals.git}{https://github.com/MohamedNagyMostafa/KITTI-MOT.Bench-Evals.git}. We highly encourage the following research to use the results to evaluate their trackers and share the tracking results.
}
\end{table*}


The tracking module performs KF for all objects stored in the memory. The process involves two central states: Update state and Prediction state. The module updates objects' trajectory estimation computed at time $t-1$ in a $2D$ camera frame $T_t^{2d}$ and $3D$ LiDAR point cloud $T_t^{3d}$ using current observation of the objects by the sensors. The module updates objects' location, velocity, and acceleration in the update state. In the prediction state, the module computes the following estimation of the objects' location in both spaces, $T^{2d}_{t+1}$ and $T^{3d}_{t+1}$, given the last updated parameters, velocity and acceleration, available for each object. In occlusion, the module only launches the prediction state, which means the estimation of objects' location and velocity will change; meanwhile, the acceleration will be the same since we use a constant acceleration model. Figure \ref{fig:occlusion} shows three scenarios for object occlusion and how our method retrieves the occluded objects given either 2D, 3D, or both trajectories information.\\

To reduce the computational time, we present the KF system and object data into matrices and apply KF estimation on the least number of points required to draw 2D and 3D bounding boxes, which are two boundary points. For the 2D bounding box, we solely employ top-left and bottom-right points to perform KF operations, and the top-left and bottom-right opposite corner points in the case of a 3D bounding box. We approach the following steps to perform KF estimation for an object: 
\begin{itemize}
\item \textit{\textbf{First-time observation}}: When the object is observed for the first time, we assign the detected bounding box location as the initial state for the object $T_{t}^{2d/3d}$, followed by the KF prediction state to obtain state estimation $T_{t+1}^{2d/3d}$. 

\item \textit{\textbf{Mono-sensor observation}}: For simplification, assume the LiDAR sensor solely observes the object. In this case, we perform both update and prediction states for KF on the 3D bounding box data and perform the prediction step on the last estimated 2D bounding box, if available.  

\item \textit{\textbf{Multi-sensor observation}}: If LiDAR and the camera observe the object, we perform the update and prediction states for KF on both 2D and 3D bounding box information. 

\item \textit{\textbf{Not observed}}: In the case of not observing an object, we still perform the prediction state for KF on both 2D and 3D bounding box information to keep track of the object for a future appearance.
\end{itemize}

\subsection{Memory Module}
\label{subsec:memory}

The memory module is essential for object occlusion. Our framework enables long-term memory to capture high-range occlusion scenarios, as shown in Figure \ref{fig:occlusion}. We discard aged objects that do not appear for subsequent frames $\epsilon$, as illustrated in Figure \ref{fig:overview}. To do so, we employ Equation \ref{eq:last_observation} to select the minimum number of subsequent frames $H_t^k$ where an object  $k$ is not observed by the camera for a number of frames $H_t^{k^{2d}}$ and LiDAR for a number of frames $H_t^{k^{3d}}$ at time $t$. 

\begin{equation}
\small
    H_t^k = min(H_t^{k^{2d}}, H_t^{k^{3d}})
    \label{eq:last_observation}
\end{equation}

As explained in section \ref{subsec:tracking}, the framework consistently updates KF estimation for all remaining objects in the memory and stores the new state estimation to be integrated into the association module later.

\section{Results and Discussion}

We split this section into four subsections. We initially introduce the utilized dataset and setup details of the conducted experiment in Section \ref{subsec:dataset}. Then, we evaluate the performance of our method with benchmark trackers under low/moderate/high detection distortion to simulate occlusion phenomena in Section \ref{subsec:detection_performance_comprison}. In the comparison, we run two of the recent benchmark tracker frameworks. In Section \ref{subsec:best_param}, we evaluate our method against recent learning and non-learning state-of-art methods. Finally, we conduct a running time comparison in Section \ref{subsec:running-time comparison}. 


\subsection{Dataset and experimental setup}
\label{subsec:dataset}

We conduct our experiment on the KITTI \cite{Geiger2012CVPR} \cite{Luiten2020IJCV} dataset that involves 21 streams with about 8,000 consecutive frames since it contains long stream duration of various urban scenarios while driving in Karlsruhe, Germany. We use LiDAR and camera information and calibration parameters provided in the dataset. We use the KITTI evaluation tool \cite{luiten2020trackeval} for evaluation. To provide a transparent comparison with benchmark models, we pick two recent benchmarks, EagerMOT \cite{kim2021eagermot} and DeepFusion-MOT \cite{wang2022deepfusionmot}, that provide modifiable source code to integrate and run various detectors. We conduct the experiments on the following hardware scheme: \textit{Processor: 11th Gen Intel Core i7-11370H 3.30GHz, GPU: NVIDIA Geforce RTX3070 Laptop GPU}. However, our framework only uses the CPU to perform tracking.

\subsection{Evaluation under detection distortion}
\label{subsec:detection_performance_comprison}

A high detection distortion means a detector frequently loses detection for objects due to poor detectors, so the distortion resembles object occlusion when we lose detection for objects in subsequent frames. In contrast, suitable detectors have minor detection distortions. In this experiment, we simulate occlusion phenomena by conducting detectors with different detection distortions: low, moderate, and high. We run the source code of two recent benchmarks, EagerMot \cite{kim2021eagermot} and DeepFusion-MOT \cite{wang2022deepfusionmot}, and use the KITTI evaluation toolkit \cite{luiten2020trackeval} \cite{Luiten2020IJCV}. We involve 20 streams from the training and validation dataset. We exclude stream \textit{0017} since we do the evaluation only for cars, and \textit{0017} does not contain any car objects. Table \ref{tab:detection_eval} shows an overview results of the experiment. We can summarize it as follows:

\textit{Poor detector \textbf{(High distortion)}}: The \textbf{1st-row} in Table \ref{tab:detection_eval} shows an experiment using poor detection performance. We conduct YOLOv3 \cite{DBLP:journals/corr/abs-1804-02767} for 2D detection and use calibration parameters to project point cloud to obtain 3D detection. Our tracker achieves $39.2\%$ and $44.5\%$ for $HOTA$ and $MOTA$, respectively, which is $3\%$ higher than EagerMOT \cite{kim2021eagermot} and $12\%$ higher than DeepFusion-MOT \cite{wang2022deepfusionmot}. Our approach has a high association accuracy of $42.8\%$, with less id switching than other trackers. Since poor detection means inconsistent detection for objects, the results reflect tracker stability under inconsistent or disconnected detection. Our tracker outperforms the two benchmarks with inconsistent detection. 

\textit{Moderate detector \textbf{(Medium distortion)}}: The \textbf{2nd-row} of Table \ref{tab:detection_eval} shows the tracking performance under a moderate performance detector. We conduct a similar setup as the previous experiment with little modifications by replacing YOLOv3 \cite{DBLP:journals/corr/abs-1804-02767} with a more sophisticated detector, RCC \cite{8099570}. We still obtain 3D detection by point cloud projection. The results show that DeepFusion-MOT \cite{wang2022deepfusionmot} cannot perform well with poor 3D detection. In contrast, EagerMOT \cite{kim2021eagermot} performance has improved by achieving $70.8\%$ for $HOTA$  and $82.2\%$ for $MOTA$. However, our tracker still outperforms EagerMOT \cite{kim2021eagermot} by nearly $11\%$ higher in both $HOTA$ and $MOTA$ metrics. Our model also has less id switching, even with poor 3D detection, which is the opposite for both EagerMOT \cite{kim2021eagermot} and DeepFusion \cite{wang2022deepfusionmot}. Finally, our model can achieve the highest $MOTA$ value, $91\%$, by conducting only one detector with $5\%$ higher than other benchmarks.

\textit{Good detector \textbf{(Low distortion)}}: The \textbf{last four rows} in Table \ref{tab:detection_eval} present tracking results utilizing sophisticated 2D and 3D detectors. In this experiment, we switch between RCC \cite{8099570} and TrackRCNN \cite{Voigtlaender19CVPR_MOTS} 2D detectors, and PointGNN \cite{Point-GNN} and PorintRCNN \cite{Shi_2019_CVPR} 3D detectors. Hence, we have four rows to represent all possible combinations. Our tracker generally has a better overall performance of $82.8\%$ and $90.7\%$ for $HOTA$ and $MOTA$, respectively. The tracker has superior performance under different combinations of 2D detectors. However, we notice that EagerMOT \cite{kim2021eagermot} has superior performance with TrackRCNN \cite{Voigtlaender19CVPR_MOTS} for 2D detection and PointGNN \cite{Point-GNN} for 3D detection. The performance of EagerMOT \cite{kim2021eagermot} with PointGNN \cite{Point-GNN} is better than PointRCNN \cite{Shi_2019_CVPR}, which explains having an almost similar performance for EagerMOT \cite{kim2021eagermot} in the sixth and fourth rows where PointGNN \cite{Point-GNN} is placed. \\

In conclusion, our tracker maintains superior performance regardless of the performance of the assigned detectors. Besides, our model has stable id switching over the experiments, which is not applied for other benchmarks when the IDSW explodes with poor and moderate detector performance. However, our method has a higher IDSW than benchmarks with good detectors for 2D and 3D, which might be caused by noise obtained by holding a long-term memory in our framework to handle prolonged occlusion.

\subsection{Evaluation with learning and non-learning benchmarks}
\label{subsec:best_param}

\begin{table}[H]

\centering

\begin{tabular}{@{}|ccccc|@{}}
     \toprule
     Method&Detector &HOTA$\uparrow$&MOTA$\uparrow$ & AMOTA$\uparrow$ \\ 
     \midrule
     PC-TCNN \cite{wu2021tracklet}& 3D &-& 89.44\% &-\\ 
     FANTrack \cite{baser2019fantrack}& 2D+3D &-&74.30\%& 40.03\%\\
     AB3DMOT \cite{weng20203d}& 3D &-& 83.35\%& 44.26\%\\
     mmMOT \cite{zhang2019robust} & 2D+3D&-& 74.07\%& 33.08\%\\
     GNN3DMOT \cite{https://doi.org/10.48550/arxiv.2006.07327}& 2D+3D &-&84.70\%&-\\
     DeepFusion-MOT \cite{wang2022deepfusionmot}&2D+3D&77.65\%&88.33\%&82.52\% \\
     EagerMOT \cite{kim2021eagermot}&2D+3D&78.53\%&87.67\%&45.62\%\\
     \midrule
     \textbf{DFR-FastMOT (Our)}& 2D &83.4\%&\textbf{93.06\%}&\textbf{90.79\%}\\
    \textbf{DFR-FastMOT (Our)}&2D+3D&\textbf{84.28\%}&91.96\%&85.36\%\\
\bottomrule

\end{tabular}

\caption{\footnotesize A comparison with other state-of-art methods includes non-learning and learning-based trackers using the KITTI evaluation dataset.}
\label{tab:other_bench}
\end{table}

We perform another evaluation with other benchmarks referring to claimed results on the evaluation dataset of KITTI and obtain tracking results of EagerMOT \cite{kim2021eagermot} and DeepFusion-MOT \cite{wang2022deepfusionmot} frameworks by running them on our machine. The comparison involves recent learning and non-learning tracking approaches. Although the PC-TCNN \cite{wu2021tracklet} tracker achieves higher performance than other benchmarks as a learning-based approach, our tracker outperforms the benchmark models, including learning-based approaches, with a $3\%$ margin in $MOTA$. As shown in Table \ref{tab:other_bench}, the tracker achieves $93.06\%$ $MOTA$ accuracy employing solely RCC 2D detector with point cloud projection. We notice that the tracker has consistent performance over the evaluation streams, allowing high $AMOTA$, $90.79\%$, dramatically higher than recent benchmarks. We achieve the highest $HOTA$, $84.28\%$, utilizing 2D RCC \cite{8099570} and 3D PointRCNN \cite{Shi_2019_CVPR} detectors. 

\subsection{Running-time comparison}
\label{subsec:running-time comparison}

To evaluate the running time, we fix 2D RCC \cite{8099570} and 3D PointRCNN \cite{Shi_2019_CVPR} detectors for all trackers, run each tracker individually on the same machine, and record time printed from the source code. We apply this experiment to all KITTI training and validation datasets to have $20$ streams with $7,763$ frames. We still exclude \textit{0017} since it contains no car objects. As shown in Table \ref{tab:total}, Our tracker processes the dataset in 1.48 seconds, which is about seven times faster than other benchmarks. The main reason for having high speed is that the method uses an algebraic model, as explained in Section \ref{subsec:association},  instead of traditional algorithmic approaches.  

\begin{table}[H]
\centering
\footnotesize
\begin{tabular}{@{}|ccc|@{}}
\toprule
Detector&Method & Time(s.) \\ 
\midrule
2D RCC \cite{8099570}&\textbf{DFR-FastMOT (Our)}&\textbf{1.48}\\
+&EagerMOT \cite{kim2021eagermot}&11.47\\
3D PorintRCNN \cite{Shi_2019_CVPR}&DeepFusion-MOT \cite{wang2022deepfusionmot}&37.38\\
\bottomrule
\end{tabular}
\caption{\footnotesize A running-time comparison for our tracker with recent benchmarks using KITTI training and evaluation datasets with $7,763$ frames in total.}

\label{tab:total}
\end{table}

\section{Conclusion}
We present a light MOT method that relies on an algebraic model to fuse and associate objects detected by a camera and LiDAR sensors. Our experiments show that the algebraic formulation for association and fusion steps dramatically reduces the computational time of the MOT method. This advantage allows long-term memory expansion in MOT methods that eventually captures complex object occlusion scenarios and boosts the overall tracking performance. We simulate object occlusion phenomena using different detection distortion levels and show that our method outperforms two of the recent benchmarks under inconsistent detection. We also evaluate our solution against learning-based and non-learning methods and show that our method outperforms recent learning-based research work by a high margin, $3\%$ in $MOTA$, and other methods by $4\%$ utilizing a mono-detector that makes it applicable for mobile robots. The work will be available at \href{https://github.com/MohamedNagyMostafa/DFR-FastMOT}{https://github.com/MohamedNagyMostafa/DFR-FastMOT}.
\bibliography{root}

\begin{thebibliography}{10}

\bibitem{chaabane2021deft}
Mohamed Chaabane, Peter Zhang, J~Ross Beveridge, and Stephen O'Hara.
\newblock Deft: Detection embeddings for tracking.
\newblock {\em arXiv preprint arXiv:2102.02267}, 2021.

\bibitem{tokmakov2021learning}
Pavel Tokmakov, Jie Li, Wolfram Burgard, and Adrien Gaidon.
\newblock Learning to track with object permanence.
\newblock In {\em Proceedings of the IEEE/CVF International Conference on
  Computer Vision}, pages 10860--10869, 2021.

\bibitem{wu2021tracklet}
Hai Wu, Qing Li, Chenglu Wen, Xin Li, Xiaoliang Fan, and Cheng Wang.
\newblock Tracklet proposal network for multi-object tracking on point clouds.
\newblock In {\em Proceedings of the International Joint Conference on
  Artificial Intelligence (IJCAI)}, pages 1165--1171, 2021.

\bibitem{wang2022deepfusionmot}
Xiyang Wang, Chunyun Fu, Zhankun Li, Ying Lai, and Jiawei He.
\newblock Deepfusionmot: A 3d multi-object tracking framework based on
  camera-lidar fusion with deep association.
\newblock {\em arXiv preprint arXiv:2202.12100}, 2022.

\bibitem{kim2021eagermot}
Aleksandr Kim, Aljo{\v{s}}a O{\v{s}}ep, and Laura Leal-Taix{\'e}.
\newblock Eagermot: 3d multi-object tracking via sensor fusion.
\newblock In {\em 2021 IEEE International Conference on Robotics and Automation
  (ICRA)}, pages 11315--11321. IEEE, 2021.

\bibitem{kf}
Rudolph~Emil Kalman.
\newblock A new approach to linear filtering and prediction problems.
\newblock {\em Transactions of the ASME--Journal of Basic Engineering},
  82(Series D):35--45, 1960.

\bibitem{shuai2021siammot}
Bing Shuai, Andrew Berneshawi, Xinyu Li, Davide Modolo, and Joseph Tighe.
\newblock Siammot: Siamese multi-object tracking.
\newblock In {\em Proceedings of the IEEE/CVF conference on computer vision and
  pattern recognition}, pages 12372--12382, 2021.

\bibitem{zheng2021improving}
Linyu Zheng, Ming Tang, Yingying Chen, Guibo Zhu, Jinqiao Wang, and Hanqing Lu.
\newblock Improving multiple object tracking with single object tracking.
\newblock In {\em Proceedings of the IEEE/CVF Conference on Computer Vision and
  Pattern Recognition}, pages 2453--2462, 2021.

\bibitem{stadler2021improving}
Daniel Stadler and Jurgen Beyerer.
\newblock Improving multiple pedestrian tracking by track management and
  occlusion handling.
\newblock In {\em Proceedings of the IEEE/CVF Conference on Computer Vision and
  Pattern Recognition}, pages 10958--10967, 2021.

\bibitem{zhou2020tracking}
Xingyi Zhou, Vladlen Koltun, and Philipp Kr{\"a}henb{\"u}hl.
\newblock Tracking objects as points.
\newblock In {\em European Conference on Computer Vision}, pages 474--490.
  Springer, 2020.

\bibitem{zhang2020multiplex}
Yang Zhang, Hao Sheng, Yubin Wu, Shuai Wang, Wei Ke, and Zhang Xiong.
\newblock Multiplex labeling graph for near-online tracking in crowded scenes.
\newblock {\em IEEE Internet of Things Journal}, 7(9):7892--7902, 2020.

\bibitem{bewley2016simple}
Alex Bewley, Zongyuan Ge, Lionel Ott, Fabio Ramos, and Ben Upcroft.
\newblock Simple online and realtime tracking.
\newblock In {\em 2016 IEEE international conference on image processing
  (ICIP)}, pages 3464--3468. IEEE, 2016.

\bibitem{kuhn1955hungarian}
H.~W. Kuhn.
\newblock The hungarian method for the assignment problem.
\newblock {\em Naval Research Logistics Quarterly}, 2(1-2):83--97, 1955.

\bibitem{bochinski2017high}
Erik Bochinski, Volker Eiselein, and Thomas Sikora.
\newblock High-speed tracking-by-detection without using image information.
\newblock In {\em 2017 14th IEEE international conference on advanced video and
  signal based surveillance (AVSS)}, pages 1--6. IEEE, 2017.

\bibitem{reich2021monocular}
Andreas Reich and Hans-Joachim Wuensche.
\newblock Monocular 3d multi-object tracking with an ekf approach for long-term
  stable tracks.
\newblock In {\em 2021 IEEE 24th International Conference on Information Fusion
  (FUSION)}, pages 1--7. IEEE, 2021.

\bibitem{luo2020fast}
Wenjie Luo, Bin Yang, and Raquel Urtasun.
\newblock Fast and furious: Real time end-to-end 3d detection, tracking and
  motion forecasting with a single convolutional net.
\newblock In {\em Proceedings of the IEEE conference on Computer Vision and
  Pattern Recognition}, pages 3569--3577, 2018.

\bibitem{wu20213d}
Hai Wu, Wenkai Han, Chenglu Wen, Xin Li, and Cheng Wang.
\newblock 3d multi-object tracking in point clouds based on prediction
  confidence-guided data association.
\newblock {\em IEEE Transactions on Intelligent Transportation Systems}, 2021.

\bibitem{unscent}
Muyuan Wang and Xiaodong Wu.
\newblock Multi-object tracking strategy of autonomous vehicle using modified
  unscented kalman filter and reference point switching.
\newblock {\em Journal of Shanghai Jiaotong University (Science)}, 26:607--614,
  10 2021.

\bibitem{extdistance}
Taeklim Kim and Tae-Hyoung Park.
\newblock Extended kalman filter (ekf) design for vehicle position tracking
  using reliability function of radar and lidar.
\newblock {\em Sensors}, 20(15), 2020.

\bibitem{hochreiter1997long}
Sepp Hochreiter and J{\"u}rgen Schmidhuber.
\newblock Long short-term memory.
\newblock {\em Neural computation}, 9(8):1735--1780, 1997.

\bibitem{yin2021centerbased}
Tianwei Yin, Xingyi Zhou, and Philipp Krahenbuhl.
\newblock Center-based 3d object detection and tracking.
\newblock In {\em Proceedings of the IEEE/CVF conference on computer vision and
  pattern recognition}, pages 11784--11793, 2021.

\bibitem{chen2019mmdetection}
Kai Chen, Jiaqi Wang, Jiangmiao Pang, Yuhang Cao, Yu~Xiong, Xiaoxiao Li,
  Shuyang Sun, Wansen Feng, Ziwei Liu, Jiarui Xu, Zheng Zhang, Dazhi Cheng,
  Chenchen Zhu, Tianheng Cheng, Qijie Zhao, Buyu Li, Xin Lu, Rui Zhu, Yue Wu,
  Jifeng Dai, Jingdong Wang, Jianping Shi, Wanli Ouyang, Chen~Change Loy, and
  Dahua Lin.
\newblock Mmdetection: Open mmlab detection toolbox and benchmark, 2019.

\bibitem{cai2017cascade}
Zhaowei Cai and Nuno Vasconcelos.
\newblock Cascade r-cnn: Delving into high quality object detection.
\newblock In {\em Proceedings of the IEEE conference on computer vision and
  pattern recognition}, pages 6154--6162, 2018.

\bibitem{8479296}
Yan-Jyun Ou, Xiang-Li Wang, Chien-Lung Huang, Jinn-Feng Jiang, Hung-Yuan Wei,
  and Kuei-Shu Hsu.
\newblock Application and simulation of cooperative driving sense systems using
  prescan software.
\newblock In {\em 2017 International Conference on Information, Communication
  and Engineering (ICICE)}, pages 173--176, 2017.

\bibitem{weng20203d}
Xinshuo Weng, Jianren Wang, David Held, and Kris Kitani.
\newblock 3d multi-object tracking: A baseline and new evaluation metrics.
\newblock In {\em 2020 IEEE/RSJ International Conference on Intelligent Robots
  and Systems (IROS)}, pages 10359--10366. IEEE, 2020.

\bibitem{object-aware}
Wentao Bao, Qi~Yu, and Yu~Kong.
\newblock Object-aware centroid voting for monocular 3d object detection.
\newblock {\em CoRR}, abs/2007.09836, 2020.

\bibitem{DBLP:journals/corr/abs-1804-02767}
Joseph Redmon and Ali Farhadi.
\newblock Yolov3: An incremental improvement.
\newblock {\em CoRR}, abs/1804.02767, 2018.

\bibitem{8099570}
Jimmy Ren, Xiaohao Chen, Jianbo Liu, Wenxiu Sun, Jiahao Pang, Qiong Yan,
  Yu-Wing Tai, and Li~Xu.
\newblock Accurate single stage detector using recurrent rolling convolution.
\newblock In {\em 2017 IEEE Conference on Computer Vision and Pattern
  Recognition (CVPR)}, pages 752--760, 2017.

\bibitem{Shi_2019_CVPR}
Shaoshuai Shi, Xiaogang Wang, and Hongsheng Li.
\newblock Pointrcnn: 3d object proposal generation and detection from point
  cloud.
\newblock In {\em The IEEE Conference on Computer Vision and Pattern
  Recognition (CVPR)}, June 2019.

\bibitem{Point-GNN}
Weijing Shi and Ragunathan~(Raj) Rajkumar.
\newblock Point-gnn: Graph neural network for 3d object detection in a point
  cloud.
\newblock In {\em The IEEE Conference on Computer Vision and Pattern
  Recognition (CVPR)}, June 2020.

\bibitem{Voigtlaender19CVPR_MOTS}
Paul Voigtlaender, Michael Krause, Aljosa Osep, Jonathon Luiten, Berin
  Balachandar~Gnana Sekar, Andreas Geiger, and Bastian Leibe.
\newblock {MOTS}: Multi-object tracking and segmentation.
\newblock In {\em CVPR}, 2019.

\bibitem{Geiger2012CVPR}
Andreas Geiger, Philip Lenz, and Raquel Urtasun.
\newblock Are we ready for autonomous driving? the kitti vision benchmark
  suite.
\newblock In {\em Conference on Computer Vision and Pattern Recognition
  (CVPR)}, 2012.

\bibitem{Luiten2020IJCV}
Jonathon Luiten, Aljosa Osep, Patrick Dendorfer, Philip Torr, Andreas Geiger,
  Laura Leal-Taixe, and Bastian Leibe.
\newblock Hota: A higher order metric for evaluating multi-object tracking.
\newblock {\em International Journal of Computer Vision (IJCV)}, 2020.

\bibitem{luiten2020trackeval}
Jonathon Luiten and Arne Hoffhues.
\newblock Trackeval.
\newblock \url{https://github.com/JonathonLuiten/TrackEval}, 2020.

\bibitem{baser2019fantrack}
Erkan Baser, Venkateshwaran Balasubramanian, Prarthana Bhattacharyya, and
  Krzysztof Czarnecki.
\newblock Fantrack: 3d multi-object tracking with feature association network.
\newblock In {\em 2019 IEEE Intelligent Vehicles Symposium (IV)}, pages
  1426--1433. IEEE, 2019.

\bibitem{zhang2019robust}
Wenwei Zhang, Hui Zhou, Shuyang Sun, Zhe Wang, Jianping Shi, and Chen~Change
  Loy.
\newblock Robust multi-modality multi-object tracking.
\newblock In {\em Proceedings of the IEEE/CVF International Conference on
  Computer Vision}, pages 2365--2374, 2019.

\bibitem{https://doi.org/10.48550/arxiv.2006.07327}
Xinshuo Weng, Yongxin Wang, Yunze Man, and Kris Kitani.
\newblock Gnn3dmot: Graph neural network for 3d multi-object tracking with
  multi-feature learning, 2020.

\end{thebibliography}

\end{document}